\documentclass[10pt,twocolumn,letterpaper]{article}

\usepackage{iccv}      
\usepackage{amsfonts}
\usepackage{amssymb}
\usepackage{amsmath}
\usepackage{multirow} 
\usepackage{colortbl}  
\usepackage{xcolor}  

\definecolor{iccvblue}{rgb}{0.21,0.49,0.74}
\usepackage[pagebackref,breaklinks,colorlinks,allcolors=iccvblue]{hyperref}

\def\paperID{5012} 
\def\confName{ICCV}
\def\confYear{2025}

\title{World4Drive: End-to-End Autonomous Driving via Intention-aware Physical Latent World Model}

\author{Yupeng Zheng$^{1,2,3}$,
Pengxuan Yang$^{1,2}$,
Zebin Xing$^{1}$,
Qichao Zhang$^{1}$\thanks{Corresponding author},
Yuhang Zheng$^{4}$,\\
Yinfeng Gao$^{1}$,
Pengfei Li$^{5}$,
Teng Zhang$^{2}$,
Zhongpu Xia$^{1}$,
Peng Jia$^{2}$,
and Dongbin Zhao$^{1}$ \\
\textsuperscript{1}CASIA,
\textsuperscript{2}Li Auto,
\textsuperscript{3}PCL,
\textsuperscript{4}NUS,
\textsuperscript{5}Tsinghua,
}

\begin{document}
\maketitle
\begin{abstract}
End-to-end autonomous driving directly generates planning trajectories from raw sensor data, yet it typically relies on costly perception supervision to extract scene information. 
A critical research challenge arises: constructing an informative driving world model to enable perception annotation-free, end-to-end planning via self-supervised learning.
In this paper, we present World4Drive, an end-to-end autonomous driving framework that employs vision foundation models to build latent world models for generating and evaluating multi-modal planning trajectories. 
Specifically, World4Drive first extracts scene features, including driving intention and world latent representations enriched with spatial-semantic priors provided by vision foundation models. 
It then generates multi-modal planning trajectories based on current scene features and driving intentions and predicts multiple intention-driven future states within the latent space.
Finally, it introduces a world model selector module to evaluate and select the best trajectory.
We achieve perception annotation-free, end-to-end planning through self-supervised alignment between actual future observations and predicted observations reconstructed from the latent space.
World4Drive achieves state-of-the-art performance without manual perception annotations on both the open-loop nuScenes and closed-loop NavSim benchmarks, demonstrating an 18.1\% relative reduction in L2 error, 46.7\% lower collision rate, and {3.75×} faster training convergence. 
Codes will be accessed at https://github.com/ucaszyp/World4Drive.
\end{abstract}    
\section{Introduction}
\label{sec:intro}

\begin{figure}[!t]
    \centering
    \includegraphics[width=0.48\textwidth]{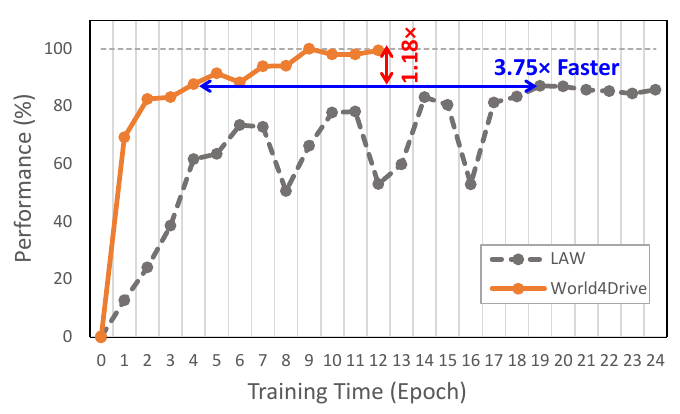}
    \vspace{-4mm}
    \caption{Our proposed World4Drive demonstrates superior convergence efficiency and performance compared to PerAct on the nuScenes dataset. As shown in the figure, where the x-axis represents training epochs (same iterations) and the y-axis shows normalized performance (calculated as the ratio of our minimum L2 error to the L2 error at each epoch). World4Drive achieves equivalent performance in 3.75$\times$ fewer training epochs and ultimately delivers a 1.18$\times$ improvement in peak performance.}
    \label{fig: teaser1}
\end{figure}

End-to-end autonomous driving integrates perception and planning into a unified, fully differentiable network. 
Given the complexity of the physical world and the uncertainty in planning intentions~\cite{chen2024vadv2}, modeling multi-modal motion planning based on a holistic understanding of physical scenes (i.e., understanding spatial, semantic, and temporal information) is a critical challenge in the field.

To enhance scene understanding, existing end-to-end approaches have explored diverse scene representations, including BEV-centric~\cite{hu2023planning,weng2024drive}, vector-based~\cite{jiang2023vad, yang2025uncad}, and sparse-centric representation~\cite{sun2024sparsedrive}. 
Some works~\cite{sima2024drivelm,marcu2024lingoqa,jin2024tod3cap} leverage multi-modal large language models to enhance scene comprehension capabilities. 
Additionally, methods like VADv2~\cite{chen2024vadv2} and Hydra-MDP~\cite{li2024hydra} model driving intentions through probabilistic planning. 
However, these approaches typically require perception annotations such as 3D bounding boxes and HD maps, which limits their scalability. 
Recently, VaVAM~\cite{bartoccioni2025vavim} leverages the learned representations of an auto-regressive video model to generate the driving trajectory directly. LAW~\cite{li2024enhancing} proposes a latent world model that constructs uni-modal latent features from raw images and acquires scene representations through temporal self-supervised learning, reducing dependence on perception annotations. 
However, extracting single-modal latent features from images struggles to capture the spatial-semantic information of the physical world and multi-modal driving intentions, resulting in slow training convergence and suboptimal performance, as shown in Fig.~\ref{fig: teaser1}.

To address these critical issues, we introduce World4Drive, an end-to-end framework that integrates multi-modal driving intentions with a latent world model to enable rational planning. This is achieved by subconsciously simulating how the physical world evolves under different driving intentions, closely mirroring the decision-making process of human drivers.

Given multi-view images and a trajectory vocabulary input, World4Drive extracts the driving intentions and world latent representations through its driving world encoding module. 
Specifically, the driving world encoding module incorporates two key components: a physical latent encoder and an intention encoder.
The physical latent encoder consists of a context encoder that leverages spatial and semantic priors from a metric depth estimation model and a vision-language model and a temporal module that aggregates temporal information to construct world latent representations enriched with physical scene context.
Concurrently, the intention encoder extracts multi-modal driving intention features from a predefined trajectory vocabulary, enabling comprehensive representation of possible driving behaviors.
Subsequently, World4Drive predicts future latent representations across multi-modal driving intentions and proposes a world model selector to select the most plausible one for self-supervised alignment training with actual world latent representations extracted from future frames. 
During inference, we fully leverage World4Drive's latent world model to evaluate and rank multi-modal trajectory candidates, enabling robust decision-making that guides the autonomous vehicle's planning process in complex driving scenarios.

World4Drive achieves state-of-the-art (SOTA) end-to-end planning performance without requiring perception annotations on the challenging nuScenes~\cite{caesar2020nuscenes} and NavSim~\cite{im2024navsim} benchmarks and is comparable to advanced perception-based models. 
Compared to LAW~\cite{li2024enhancing}, a previous SOTA unsupervised method, our method significantly reduces average planning displacement error by 18.2\% (from 0.61m to 0.50m) and average collision rate by 46.7\% (from 0.30\% to 0.16\%). 
More remarkably, our approach achieves more than three times faster convergence speed by appropriately incorporating spatial-semantic priors from vision foundation models.

Our key contributions are summarized as follows: 
\begin{itemize}
    \item Inspired by human driver decision processes, we propose an intention-aware latent world model that innovatively uses the world model to generate and evaluate multi-modal trajectories under different intentions.
    \item To enhance the world model's understanding of the physical world without perception annotations, we design a novel driving world encoding module that leverages prior knowledge of vision foundation models to extract physical latent representations of the driving environment. 
    \item Our method achieves SOTA planning performance without perception annotations on open-loop nuScenes and closed-loop NavSim benchmarks while significantly accelerating convergence speed.
\end{itemize}

\section{Related Works}
\label{sec:rel}
\subsection{End-to-End Autonomous Driving}
In recent years, with advancements in perception technologies using BEV-centric~\cite{li2024bevformer, yang2023bevformer}, vector-based~\cite{liao2022maptr}, and sparse-centric~\cite{lin2022sparse4d} scene representations, vision-based end-to-end autonomous driving has garnered increasing attention.
Models such as UniAD~\cite{hu2023planning}, VAD~\cite{jiang2023vad}, and SparseDrive~\cite{sun2024sparsedrive} have explored those diverse representations, establishing end-to-end architectures including perception, prediction, and planning. 
GenAD~\cite{zheng2024genad} leverages generative models to produce trajectories, while some methods~\cite{weng2024drive, zheng2024preliminary,jia2025drivetransformer} implement parallelized end-to-end structures. 
To consider the intention uncertainty in planning, VADv2~\cite{chen2024vadv2} and Hydra-MDP~\cite{li2024hydra} model driving intentions through probabilistic planning. 
DiffusionDrive and GoalFlow~\cite{liao2024diffusiondrive, xing2025goalflow} explores end-to-end methodologies based on diffusion models~\cite{ho2020denoising}. 
Furthermore, with the evolution of large language models (LLMs), several approaches~\cite{marcu2024lingoqa, sima2024drivelm} enhance scene information through language models. 
For instance, VLP~\cite{pan2024vlp} incorporates linguistic understanding into scene information via contrastive learning. 
DriveVLM~\cite{tian2024drivevlm} constructs a dual slow-fast system that integrates vision-language model capabilities into the decision space. 
TOKEN~\cite{tian2024tokenize} utilizes large language models to enhance object-level perception, improving planning capabilities in long-tail scenarios. 
These methods typically require extensive and costly perception or QAs annotations, which limits their scalability.

 \begin{figure*}[ht]
    \centering
    \includegraphics[width=1\textwidth]{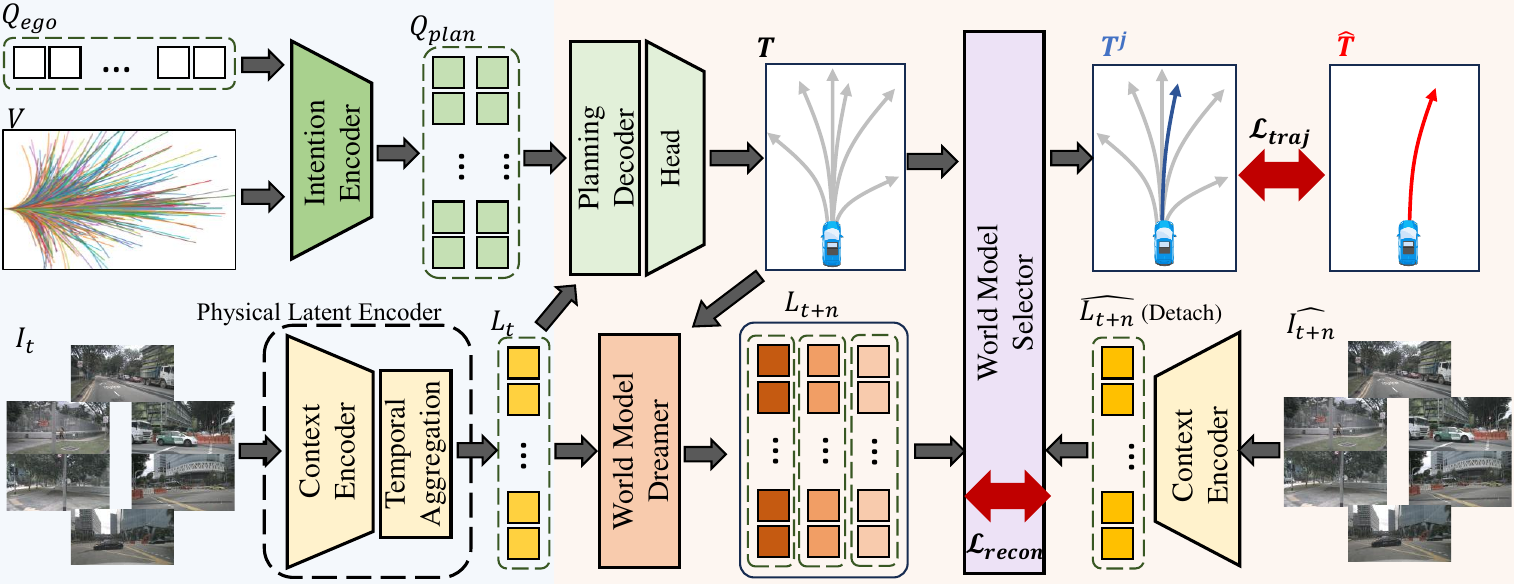}
    \caption{We propose World4Drive, a novel approach that constructs an intention-aware latent world model to generate, evaluate, and rank multi-modal trajectories under multi-modal driving intentions.}
    \label{fig:teaser}
\end{figure*}

\subsection{Autonomous Driving World Models}
World models in autonomous driving aim to predict scene evolution following various actions. 
These include image-based video generation, 3D world models based on representations such as point clouds and occupancy grids, and latent feature-based future world generation. 
Image-based video generation encompasses driving video approaches using diffusion models like DriveDreamer~\cite{wang2024drivedreamer,zhao2024drivedreamer}, Vista~\cite{gao2024vista}, and Drive-WM~\cite{wang2024driving}, as well as driving video generation methods based on autoregressive models such as DriveWorld~\cite{min2024driveworld} and GAIA\cite{hu2023gaia}. 
3D world models include point cloud-based world models~\cite{zhangcopilot4d} and occupancy-based world models~\cite{zheng2024occworld, gu2024dome}. These construct models in 3D space to better capture dynamic changes in 3D scenes. 
 Recently, approaches like VaVAM~\cite{bartoccioni2025vavim} and LAW~\cite{li2024enhancing} have employed video generation techniques to learn scene representations through self-supervised learning, eliminating the dependency on perception annotations. Specifically, LAW proposes latent world models that predict a single future scene latent feature through self-supervision learning, achieves SOTA end-to-end planning performance. {However, constructing single-modal latent features from raw images often struggles to capture the spatial-semantic scene information and the uncertainty of multi-modal driving intentions, resulting in suboptimal performance.} 

\section{Method}
\label{sec:method}

\subsection{Overview}
The overall pipeline of World4Drive is illustrated in Fig.~\ref{fig:teaser}. 
World4Drive comprises two key modules: 1) Driving World Encoding (Sec.~\ref{sec: DWE}), which extracts driving intention and physical world latent representations from RGB images and trajectory vocabulary, and
2) Intention-aware World Model (Sec.~\ref{sec: IWE}), which predicts the latent representation of the future world according to multi-modal driving intentions and scores multi-modal planning trajectories via the world model selector.
The two key modules are tightly coupled, enabling autonomous driving vehicles to imagine the future world under various intentions while achieving vision-based end-to-end planning without requiring perception annotations.

\subsection{Driving World Encoding}
\label{sec: DWE}
In the Driving World Encoding module, we introduce an intention encoder that takes vocabulary as input to extract driving intentions and a physical latent encoder that utilizes a vision-language model and  a metric depth estimation model to extract world latent representations that are aware of spatial, semantic and temporal context.

\subsubsection{Intention Encoder}
\label{sec: IE}
Given a randomly initialized ego query $Q_{ego}$ and a trajectory vocabulary $\mathcal{V} \in \mathbb{R}^{N \times S \times 2}$ input~\cite{chen2024vadv2}, we first obtain the intention point $P_I \in \mathbb{R}^{3 \times K \times 2}$ by adopting the k-means clustering algorithm on the endpoints of $\mathcal{V}$. 
Among them, $N$ represents the number of trajectories in the trajectory vocabulary, 3 represents the three commands type(e.g., left, right, straight), $K$ represents the number of intentions for each command type, and $S$ represents the number of waypoints in each trajectory.
Then we obtain the intention query $Q_I$ with sinusoidal position encoding. 
Finally, we utilize a self-attention layer to obtain the intention-aware multi-modal planning query $Q_{plan}$. Formally,
\begin{equation}
Q_{plan} = {\rm SelfAttention}((Q_{ego} + Q_{I})).
\end{equation}
By default, $N$ is set to 8192, and $K$ is set to 6.

\subsubsection{Physical World Latent Encoding}
\label{sec: PWE}
\begin{figure}[!t]
    \centering
    \includegraphics[width=0.48\textwidth]{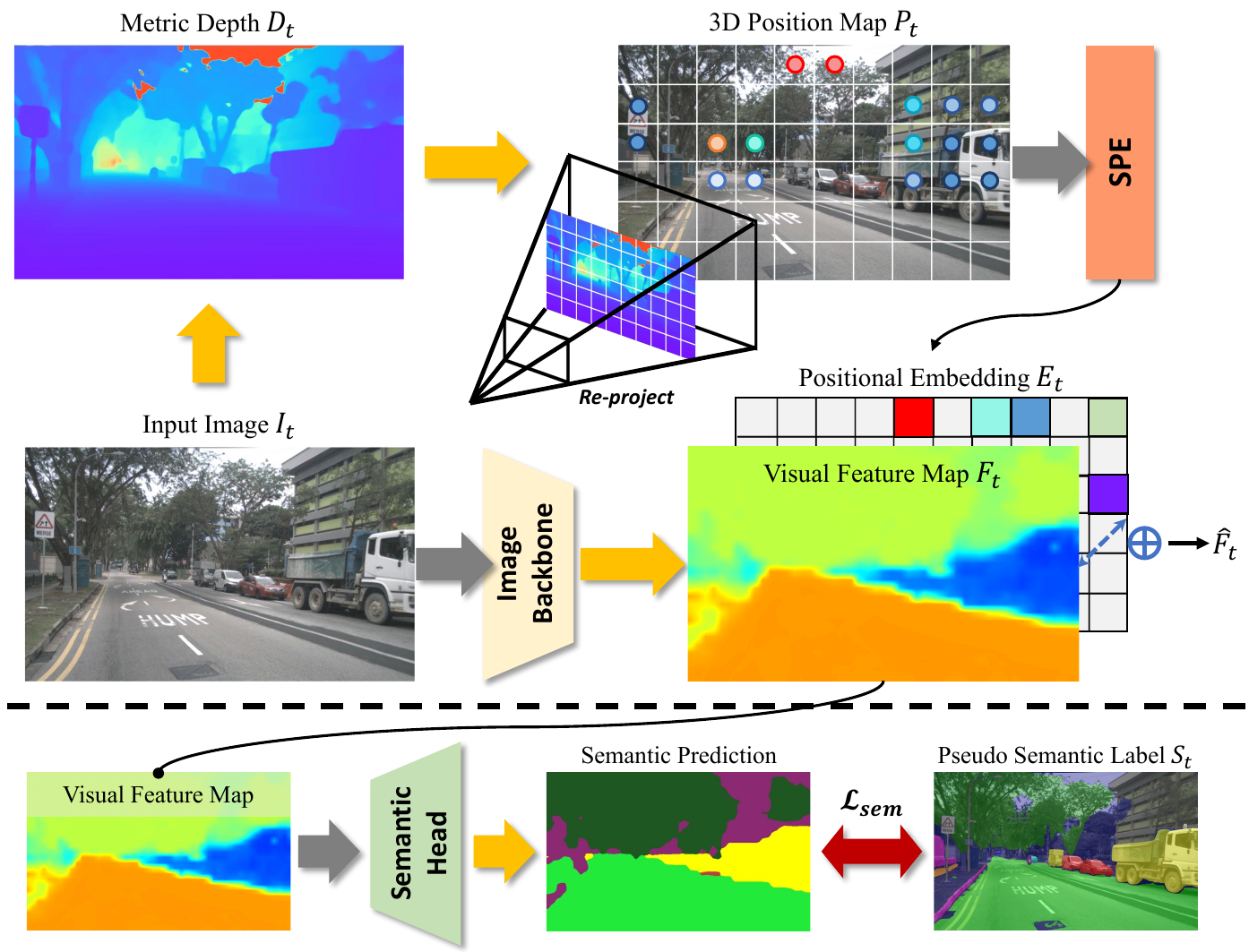}
    \vspace{-4mm}
    \caption{Detailed pipeline of context encoder. It consists of a 3D spatial encoding module and a semantic understanding module, achieving a holistic understanding of the physical world.}
    \vspace{-2mm}
    \label{fig: context}
\end{figure}
We introduce the Physical World Latent Encoding module to extract the world latent representations with a holistic understanding (i.e., spatial and semantic perception capabilities) of the 3D physical world. 
The module consists of a context encoder (see Fig.~\ref{fig: context} for details) to incorporate spatial and semantic prior and a temporal aggregation module to enhance temporal context.

\textbf{Context Encoder. }
Given a frame of multi-view images $I_t \in \mathbb{R}^{M \times H \times W \times 3}$ at timestep $t$ as input, we first extract the corresponding image features $F_t \in \mathbb{R}^{M\times h \times w \times D}$ with an image backbone, where $D$ is the feature dimension and $M$ represents the number of camera views. 
Previous work, LAW~\cite{li2024enhancing}, directly extracts camera features as world latent representation, lacking spatial and semantic understanding of driving scenarios. To address this issue, we introduce spatial-semantic priors with open-vocabulary semantic supervision and 3D geometry-aware positional encoding.

\noindent\textbf{Semantic Understanding.}
We utilize the vision-language model Grouded-SAM~\cite{ren2024grounded} to produce pseudo semantic labels. Given the prompt for the object of interest, we obtain 2D bounding boxes and the corresponding semantic mask $S_t \in \mathbb{R}^{H \times W \times C}$ via the Grouded-SAM model. Formally,
\begin{equation}
    S_t = {\rm GroundedSAM}(F_t),
\end{equation}
we only keep labels with high confidence to reduce incorrect labeling.
Finally, we leverage cross-entropy loss $\mathcal{L}_{sem}$ to enhance the semantic understanding of the latent representations.

\noindent\textbf{3D Spatial Encoding.}
3D Spatial Encoding aims to provide the model with accurate positional information in the physical world. 
Previous work, PETR~\cite{liu2022petr}, utilizes a post-projection approach by generating 3D meshgrids to provide different 3D positional encodings for each pixel. 
Inspired by this concept, we provide the scale-aware depth for each pixel to represent 3D space, offering accurate spatial understanding for end-to-end planning.
Specifically, we employ a metric depth model~\cite{hu2024metric3d, yin2023metric3d} to estimate multi-view depth maps $D_t \in \mathbb{R}^{M \times h \times w}$. 
In contrast to PETR, we adopt a forward projection approach, obtaining the 3D position in the ego coordinate system $p = \{x, y, z\}$ of each pixel $(u, v)$ through depth maps and the camera intrinsic matrix.
Thus, we can generate 3D position maps $P_t \in \mathbb{R}^{M \times h \times w \times 3} $. 
Subsequently, we encode these 3D positions using sinusoidal positional encoding and obtain corresponding positional embedding $E_t \in \mathbb{R}^{M \times h \times w \times D} $ through a learnable MLP. Formally,
\begin{equation}
    E_t = {\rm MLP}({\rm SPE}(P_t)),
\end{equation}
where {\rm SPE(·)} is the sinusoidal position encoding.
Finally, by adding positional embedding $E_t$ to the image features $F_t$, we obtain the semantic-spatial-aware visual feature $\hat{F_t}$.

\textbf{Temporal Aggregation.}
Different from previous work~\cite{li2024enhancing} that uses randomly initialized queries to obtain latent representation.
We employ a temporal aggregation module to obtain latent representation enriched with the temporal context.
In particular, we preserve the visual feature $\hat{F}_{t-1}$ at the prior timestamp $t-1$.  
To enhance the temporal information in the world latent representation, we aggregate historical information into the current visual features through cross-attention mechanisms to obtain world latent representations $L_t \in \mathbb{R}^{M \times h \times w \times D} $. Formally,
\begin{equation}
    L_t = {\rm CrossAttention}(\hat{F_t}, \hat{F}_{t-1}).
\end{equation}

The proposed Physical World Latent Encoder enriches the world latent representations with spatial, semantic, and temporal information, providing a holistic understanding of the dynamic driving environment, which is crucial for imagining the future world.

\subsection{Planning with Intention-aware World Model}
\label{sec: IWE}

In this section, we propose the intention-aware world model to predict the latent representation of the future world according to multi-modal driving intentions (Sec.~\ref{sec: WMD}) and score multi-modal planning trajectories via the world model selector (Sec.~\ref{sec: WMS}).

\subsubsection{Intention-aware World Model Dreamer}
\label{sec: WMD}
\textbf{Action Encoding.}
Given the intention-aware multi-modal planning query $Q_{plan}$, we first employ a cross-attention layer to aggregate scene context into $Q_{plan}$.
Then, we obtain the multi-modal trajectories $T = \{T^1,...,T^K\} \in \mathbb{R}^{K\times S \times 2}$ with an MLP layer.
Formally,
\begin{equation}
    T = {\rm MLP}({\rm CrossAttention}(Q_{plan}, L_t)).
\end{equation}
Finally, we employ an action encoder (an MLP layer) to acquire intention-aware action tokens $A \in \mathbb{R}^{K \times D}$, where $K$ is the number of intentions.

\noindent\textbf{Intention-aware World Model Prediction. }
Our objective is to predict future world latents $L_{t+n} = \{L_{t+n}^1,...,L_{t+n}^K\}$  following actions corresponding to each driving intention, where $n$ is the timestamp interval.
We concatenate action tokens $A$ and world latent $L$ along the dimension channel. 
Different from the previous work, we randomly initialize a learnable query $Q_{future}$ and adopt multi-layer cross-attention as the predictor. Formally,
\begin{equation}
    L_{t+n} = {\rm CrossAttention}(Q_{future}, {\rm Concat}(A, L)).
\end{equation}
By default, we set $n=3$, the ablation study of timestamp interval $n$ can be seen in the supplementary material.
\subsubsection{World Model Selector}
\label{sec: WMS}
\begin{figure}[!t]
    \centering
    \includegraphics[width=0.48\textwidth]{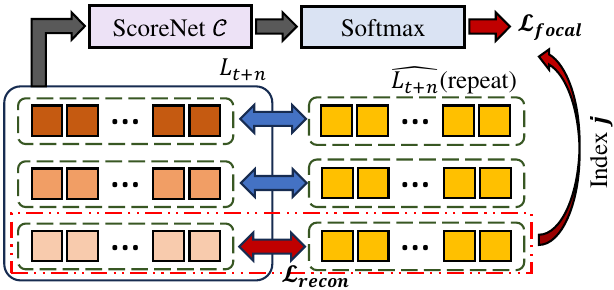}
    \vspace{-4mm}
    \caption{Detailed pipeline of World Model Selector. By computing and comparing feature distances between predicted latents and actual latent, the model selects the most plausible latent and its corresponding trajectory as output. The selector employs reconstruction loss between the selected latent and the actual latent to self-supervised learn scene representations while simultaneously training the world model scoring network using focal loss between predicted scores and the index of the selected latent.}
    \vspace{-2mm}
    \label{fig: wms}
\end{figure}

We propose a World Model Selector module that evaluates trajectories under $K$ different intentions through the latent world model and selects the reasonable trajectory from among them. 
The detailed architecture is illustrated in Fig.~\ref{fig: wms}.
In particular, given the predicted intention-aware future latent $L_{t+n}$ and the actual future latent $\hat{L_{t+n}}$, we compute the feature distance between the predicted latent representation and the actual latent representation for each modality. 
The modality with the minimum distance is selected as the final selected modality (assume the index of this modality is $j$), the corresponding latent distance is used as the reconstruction loss $\mathcal{L}_{recon}$ for optimization, and the corresponding trajectory $T^j$ is selected as the final planning trajectory.
Simultaneously, we employ a classification network as the ScoreNet, $\mathcal{C}$, to predict scores $\mathbb{S} = \{\mathbb{S}^1, ...\mathbb{S}^K\}$ corresponding to the $K$ modalities. Formally,
\begin{equation}
    \mathbb{S} = {\rm Softmax}(\mathcal{C}(L_{t+n}))
\end{equation}
We utilize a focal loss between scores $\mathbb{S}$ and the selected modality index $j$ to optimize the world model scoring network.

Notably, 1) during inference, we directly select the trajectory corresponding to the world model with the highest score as the final trajectory, and 2) we employ the MSE loss to compute the latent distance. We do ablation on other losses in the supplementary material.

\subsection{Training Loss}
Following previous works, we apply the $L_1$ loss $\mathcal{L}_{traj}$ to guide the final planning trajectory $T^j$ with the expert trajectory $\hat{T}$.
World4Drive is end-to-end trainable. Therefore, the final loss for end-to-end training is
\begin{equation}
    \mathcal{L} = \alpha\mathcal{L}_{sem} + \beta\mathcal{L}_{recon} + \gamma\mathcal{L}_{score} + \eta\mathcal{L}_{traj},
\end{equation}
we set $\alpha=0.2, \beta=0.2, \gamma=0.5, \eta=1.0$ by default.
\section{Experiments}
\subsection{Benchmarks}
\textbf{Open-loop nuScenes Benchmark.}
The nuScenes~\cite{caesar2020nuscenes} benchmark is an open-loop evaluation framework developed for the real-world nuScenes dataset. The nuScenes dataset comprises 1000 driving videos captured across diverse environments. In line with previous methodologies~\cite{hu2023planning, jiang2023vad}, we employ displacement error (L2) and collision rate (CR) as evaluation metrics for the predicted trajectories, which are sampled at 2Hz over a 3-second horizon.

\noindent\textbf{Closed-loop NavSim Benchmark.}
The NavSim~\cite{im2024navsim} benchmark is built upon the OpenScene~\cite{openscene2023} dataset, encompassing 1192 training scenarios and 136 test scenarios, with a total of over 100,000 keyframes. In alignment with the officially provided baseline, we interpolate the predicted trajectories (sampled at 2Hz over a 4-second horizon) using an LQR controller. The model performance is evaluated through closed-loop PDM scores (PDMS), which are calculated based on five key factors: no at-fault collision (NC), drivable area compliance (DAC), time-to-collision (TTC), comfort (Comf.), and ego progress (EP).

\begin{table*}[htb]
    \centering
    \small
    \setlength{\tabcolsep}{0.022\linewidth}
    \renewcommand{\arraystretch}{1.0} 
    \caption{End-to-end planning results on nuScenes benchmark~\cite{caesar2020nuscenes}}
    \vspace{-2mm}
    \resizebox{0.9\textwidth}{!}{ 
    \begin{tabular}{l|ccc>{\columncolor[gray]{0.9}}c|ccc>{\columncolor[gray]{0.9}}c}
        \toprule
        \multirow{2}{*}{\text{Method}} & \multicolumn{4}{c|}{L2 (m) $\downarrow$} & \multicolumn{4}{c}{\text{Collision Rate (\%) $\downarrow$}}  \\
        & \text{1s} & \text{2s} & \text{3s} & \text{Avg.} & \text{1s} & \text{2s} & \text{3s} & \text{Avg.} \\
        \midrule
        \cellcolor{blue!30} ST-P3~\cite{hu2022st} & 1.33 & 2.11 & 2.90 & 2.11 & 0.23 & 0.62 & 1.27 & 0.71  \\
        \cellcolor{blue!30} OccNet~\cite{tong2023scene} & 1.29 & 2.13 & 2.99 & 2.13 & 0.21 & 0.59 & 1.37 & 0.72 \\
        \cellcolor{blue!30} UniAD~\cite{hu2023planning} & 0.48 & 0.96 & 1.65 & 1.03 & \underline{0.05} & 0.17 & 0.71 & 0.31  \\
        \cellcolor{blue!30} VAD~\cite{jiang2023vad} & 0.41 & 0.70 & 1.05 & 0.72 & 0.07 & 0.18 & 0.43 & 0.23  \\
        \cellcolor{blue!30} PPAD~\cite{chen2024ppad} & 0.31 & 0.56 & 0.87 & 0.58 & 0.08 & \underline{0.12} & 0.38 & 0.19 \\
        \cellcolor{blue!30} GenAD~\cite{zheng2024genad} & 0.28 & 0.49 & 0.78 & 0.52 & 0.08 & 0.14 & 0.34 & 0.19 \\
        \cellcolor{blue!30} LAW*~\cite{li2024enhancing} (Perception-based) & \underline{0.24} & \textbf{0.46} & \textbf{0.76} & \textbf{0.49} & 0.08 & \textbf{0.10} & 0.39 & 0.19  \\
        \midrule
        \cellcolor{red!30} BEV-Planner~\cite{li2024ego} & 0.30 & 0.52 & 0.83 & 0.55 & 0.10 & 0.37 & 1.30 & 0.59 \\
        \cellcolor{red!30} LAW*~\cite{li2024enhancing} (Perception-free) & 0.26 & 0.57 & 1.01 & 0.61 & 0.14 & 0.21 & 0.54 & 0.30 \\
        \cellcolor{red!30} World4Drive (Ours) & \textbf{0.23} & \underline{0.47} & \underline{0.81} & \underline{0.50} & \textbf{0.02} &\underline{0.12} & \textbf{0.33} & \textbf{0.16}  \\
        \bottomrule
        \multicolumn{9}{l}{* LAW~\cite{li2024enhancing} adopt Swin-Tiny~\cite{liu2021swin} as image backbone while other methods adopt ResNet-50~\cite{he2016deep} as image backbone.}\\

    \end{tabular}
    }
    \label{tab1}

   \vspace{1mm}
\end{table*}

\begin{table*}[!ht]
    \centering
    \small
    \setlength{\tabcolsep}{0.015\linewidth}
    \renewcommand{\arraystretch}{1.0} 
    \caption{End-to-end planning results on NavSim benchmark~\cite{caesar2020nuscenes}}
    \vspace{-2mm}
    \resizebox{0.9\textwidth}{!}{ 
\begin{tabular}{l|ccr|cccccc}
    \toprule
    Method & Input & NC $\uparrow$ &DAC $\uparrow$ & TTC $\uparrow$& Comf. $\uparrow$ & EP $\uparrow$ & \cellcolor{gray!30}PDMS $\uparrow$  \\
    \midrule
    \cellcolor{blue!30} UniAD~\cite{hu2023planning} & C & 97.8 & 91.9 & 92.9 & \textbf{100.0} & 78.8 & \cellcolor{gray!30}83.4 \\
    \cellcolor{blue!30} PARA-Drive~\cite{sun2024sparsedrive} & C & \underline{97.9} & 92.4 & \underline{93.0} & 99.8 & 79.3 & \cellcolor{gray!30}84.0 \\
    \cellcolor{blue!30} LTF~\cite{prakash2021multi} & C & 97.4 & 92.8 & 92.4 & \textbf{100.0} & 79.0 & \cellcolor{gray!30}83.8 \\
    \cellcolor{blue!30} Transfuser~\cite{prakash2021multi} & C \& L & 97.7 & 92.8 & 92.8 & \textbf{100.0} & 79.2 & \cellcolor{gray!30}84.0 \\
    \cellcolor{blue!30} VADv2~\cite{chen2024vadv2} & C \& L & 97.2 & 89.1 & 91.6 & \textbf{100.0} & 76.0 & \cellcolor{gray!30}80.9 \\
    \cellcolor{blue!30} Hydra-MDP~\cite{li2024hydra} & C \& L & \underline{97.9} & 91.7 & 92.9 & \textbf{100.0} & 77.6 & \cellcolor{gray!30}83.0 \\
    \cellcolor{blue!30} DiffusionDrive~\cite{liao2024diffusiondrive} & C \& L & \textbf{98.2}  & \textbf{96.2}  & \textbf{94.7}  & \textbf{100.0}  & \textbf{82.2}  & \cellcolor{gray!30}\textbf{88.1}\\
    \midrule
    \cellcolor{red!30} Ego-MLP & E & 93.0  & 77.3  & 83.6  & \textbf{100.0}  & {62.8} & \cellcolor{gray!30}{65.6} \\
    \cellcolor{red!30} LAW (Perception-free)~\cite{li2024enhancing} & C & 97.2  & {93.3}  & 91.9  & \textbf{100.0}  & {78.8}  & \cellcolor{gray!30}{83.8}\\    
    \cellcolor{red!30} World4Drive (Ours) & C & {97.4}  & \underline{94.3}  & {92.8}  & \textbf{100.0}  & \underline{79.9}  & \cellcolor{gray!30}\underline{85.1}\\
    \bottomrule
    \multicolumn{8}{l}{In the Input column, C represents camera modality, L represents lidar modality, and E represents ego status.}\\
\end{tabular}%
}
\label{tab2}
\end{table*}

\subsection{Implementation Details}
\textbf{nuScenes Benchmark.}
Following the VAD-Tiny~\cite{jiang2023vad} configuration, we employ ResNet-50~\cite{he2016deep} as the image backbone, processing 6 surround-view images with a resolution of $360 \times 640$. The driving commands align with previous works~\cite{hu2023planning}, comprising three types: left, right, and straight. For each command, we predict 6 planning trajectories and select the one with the highest score corresponding to the world model as the final planning trajectory. We train the model for 12 epochs on 8 NVIDIA 3090 GPUs with a total batch size of 8 and an initial learning rate of 5e-5.

\noindent\textbf{NavSim Benchmark.}
Consistent with the NavSim benchmark, our closed-loop model takes a concatenated image formed by stitching front, front-left, and front-right camera views as input, which is then resized to $256 \times 1024$. We employ ResNet-34~\cite{he2016deep} to extract image features. We train the model for 60 epochs on 8 NVIDIA 3090 GPUs with a total batch size of 64. Since LAW~\cite{li2024enhancing} is not open-sourced, we reimplement and evaluate it under settings identical to ours.
For vision foundation models, we utilize the giant model from Metric3D v2~\cite{hu2024metric3d} for depth estimation and Grounded-SAM~\cite{ren2024grounded} for semantic segmentation.

\subsection{Main Results}
As demonstrated in Tab.~\ref{tab1}, we compare our proposed framework with several SOTA methods. 
Methods highlighted with \textcolor{blue!60}{blue background} in the table require manual perception annotations, whereas methods highlighted with \textcolor{red!60}{red background} do not require manual perception annotations for training and inference. 
World4Drive achieves SOTA performance among perception annotation-free approaches, demonstrating an 18.0\% reduction in L2 error and a 46.7\% reduction in collision rate compared to strong baselines.
Furthermore, World4Drive achieves the lowest collision rate among all methods. Compared to LAW, the SOTA perception-based approach, our method shows only a modest increase of less than 2\% in L2 error while significantly improving safety metrics.

As demonstrated in Tab.~\ref{tab2}, World4Drive also achieves competitive performance in closed-loop metric PDMS.
Compared to the baseline, our approach demonstrates significant improvements in Time-to-Collision (TTC) and Drivable Area Compliance (DAC) metrics. 
These metrics specifically evaluate an autonomous vehicle's spatial awareness and understanding of the drivable area.
The result indicates that incorporating the vision foundation model priors substantially enhances the model’s comprehensive understanding of the physical world.
Moreover, our closed-loop metrics surpass those of other methods requiring perception annotations, with the exception of DiffusionDrive~\cite{liao2024diffusiondrive}.

\subsection{Ablation Study}
In this section, we conduct several ablation studies to explore the effectiveness, robustness, and stability of our proposed World4Drive. All of the ablation experiments are conducted on the nuScenes~\cite{caesar2020nuscenes} benchmark. 
All L2 errors and collision rates are averaged over a 3-second prediction horizon.

\subsubsection{Effectiveness of Each Component}
In this section, we assess the effectiveness of each component in our method. Detailed results are demonstrated in Tab.~\ref{tab: ablation}.
Row 1 demonstrates the result of our baseline, LAW, which only has a single-modal world model. 
Comparing row 1 and row 2, we observe that incorporating vehicle intention significantly reduces L2 error and collision rate. 
Further, the comparison between row 1 and row 4 demonstrates substantial planning performance improvements when integrating priors from vision foundation models and vision language models, highlighting the critical importance of comprehensive physical world understanding.

To explore the contribution of different perceptual components, we conducted a more detailed analysis. 
Comparing rows 2 and 3, we find that introducing spatial priors enhances positional awareness, thereby improving trajectory-fitting capabilities. 
Similarly, a comparison between rows 3 and 6 reveals that semantic priors significantly reduce collision rates, suggesting better obstacle understanding.

Finally, we investigate the necessity of combining intentions with world modeling. 
The comparison between rows 4 and 6 demonstrates that adding intention modeling substantially improves planning quality, which is intuitive as intentions provide multiple planning possibilities, enabling the model to select safer trajectories. However, the comparison between rows 5 and 6 reveals that intentions alone, without world modeling, actually lead to degraded planning performance.
This confirms the crucial role of the world model in evaluating and ranking multi-modal intentions.

\begin{table}[t]
    \centering
    \setlength{\tabcolsep}{0.018\linewidth} %
    \renewcommand{\arraystretch}{1.5} %
    \caption{Ablation study of each proposed component}
    \vspace{-2mm}
    \resizebox{0.47\textwidth}{!}{
    \begin{tabular}{c|cc|cc|c|c}
        \toprule
        \multirow{2}{*}{ID} & \multicolumn{2}{c|}{Physical Latent Encoder} &  \multicolumn{2}{c|}{Intention-aware WM} & \multirow{2}{*}{L2} &  \multirow{2}{*}{Collision} \\
        \cline{2-5}
        & Depth& Semantic & WM & Intentions & &  \\ 

        \midrule
        1 & & & \checkmark &  & 0.61 & 0.30\\
        2 & & & \checkmark &  \checkmark &0.55& 0.25\\
        3 & \checkmark & & \checkmark &  \checkmark &0.51& 0.29\\
        4& \checkmark & \checkmark & \checkmark &  &0.49& 0.26\\
        5& \checkmark & \checkmark & & \checkmark & 0.61 & 0.36 \\
        6& \checkmark & \checkmark & \checkmark & \checkmark & 0.50 & 0.16\\
        \bottomrule
    \end{tabular}
    }
    \label{tab: ablation}
\end{table}

\subsubsection{Performance In Different Driving Conditions}
\begin{table}[t]
    \centering
    \setlength{\tabcolsep}{5pt} 
    \renewcommand{\arraystretch}{1.2} 
    \caption{Performance under different driving conditions.}
    \vspace{-2mm}
    \resizebox{0.47\textwidth}{!}{ 
    \begin{tabular}{cc|cc|cc}
        \toprule
        \multirow{2}{*}{\text{ID}} & \multirow{2}{*}{\text{Condition}} & \multicolumn{2}{c|}{World4Drive (Ours)} & \multicolumn{2}{c}{LAW~\cite{li2024enhancing}}  \\
         & & {L2 (m) $\downarrow$} & {\text{Collsion (\%) $\downarrow$}} & {L2 (m) $\downarrow$} & {\text{Collsion (\%) $\downarrow$}} \\
         \midrule
        1 & All & \textbf{0.50} & \textbf{0.16} & 0.61 & 0.30 \\
        2 & Day & \textbf{0.47} & \textbf{0.16} & 0.56 & 0.26 \\
        3 & Night & 0.76 & \textbf{0.08} & \textbf{0.67} & 0.22 \\
        4 & Sunny & \textbf{0.50} & \textbf{0.18} & 0.58 & 0.29 \\
        5 & Rainy & \textbf{0.49} & \textbf{0.05} & 0.54 & 0.16\\
        \midrule
    \end{tabular}
    }
    \label{tab: condition}
\end{table}
In this section, we analyze planning performance across diverse driving conditions, including varying weather conditions, illumination settings, and driving maneuvers. 
Following the official nuScenes scene descriptions, we categorize weather as sunny or rainy, illumination as day or night, and driving maneuvers as left, straight driving, or right.

Tab.~\ref{tab: condition} presents a comparative analysis of our method against the baseline, LAW~\cite{li2024enhancing}, across different weather and lighting conditions. 
Our approach consistently outperforms LAW across almost all environmental scenarios. Notably, compared to LAW, in challenging nighttime and rainy conditions, our method reduces collision rates by 63.7\% and 68.8\%, respectively. 
This significant improvement can be attributed to the integration of priors from vision foundation models, which enables our system to comprehend higher-dimensional spatial and semantic information about the physical environment. 
Consequently, our approach demonstrates greater robustness to photometric inconsistencies inherent in nighttime and rainy weather conditions, which typically impede the temporal self-supervised training of latent world models in baseline methods.

Tab.~\ref{tab: maneuvers} demonstrates the comparison of planning performance under different driving maneuvers between our method and LAW.
Compared to LAW, our method generates significantly safer planning trajectories across a variety of driving maneuvers.

The superior planning performance under diverse driving conditions demonstrates the effectiveness and robustness of our approach.

\begin{table}[htb]
    \centering
    \setlength{\tabcolsep}{0.013\linewidth} 
    \renewcommand{\arraystretch}{1.2} 
    \caption{Performance under driving maneuvers.}
    \vspace{-2mm}
    \resizebox{0.49\textwidth}{!}{ 
    \begin{tabular}{l|ccc>{\columncolor[gray]{0.9}}c|ccc>{\columncolor[gray]{0.9}}c}
        \toprule
        \multirow{2}{*}{Model} & \multicolumn{4}{c|}{L2 (m) $\downarrow$} & \multicolumn{4}{c}{\text{Collsion (\%) $\downarrow$}}  \\
        &  Left &  Right & Straight & All &  Left&  Right & Straight & All \\
        \midrule
        LAW & 0.67& 0.71& 0.58 & 0.61 & 0.54 & 0.23 & 0.29 & 0.30 \\
        World4Drive& \textbf{0.63} & \textbf{0.69} & \textbf{0.48} & \textbf{0.50} & \textbf{0.40} & \textbf{0.20} & \textbf{0.14} & \textbf{0.16} \\
        \midrule
    \end{tabular}
    }
    \label{tab: maneuvers}
\end{table}

\subsubsection{Scability of World4Drive}
\begin{table}[t]
    \centering
    \setlength{\tabcolsep}{0.025\linewidth} %
    \renewcommand{\arraystretch}{1.0} %
    \caption{Ablation study of scalability}
    \vspace{-1.5mm}
    \resizebox{0.46\textwidth}{!}{
    \begin{tabular}{c|cc|c|c}
        \toprule
        ID & Backbone & Dimension & L2 (m) $\downarrow$ & Collsion (\%) $\downarrow$ \\ 
        \midrule
        1 & ResNet-34 & 256 & 0.52 & 0.25 \\
        2 & ResNet-50 & 128 & 0.55 & 0.27 \\
        3 & ResNet-50 & 256 & 0.50 & 0.16 \\
        4 & ResNet-50 & 384 & 0.49 & 0.10 \\
        5 & ResNet-101 & 256 & 0.47 & 0.14 \\
        \bottomrule
    \end{tabular}
    }
    \vspace{-2mm}
    \label{tab: scale}
\end{table}

To explore the scalability of our approach, we conduct experiments by varying both the size of the hidden dimension $D$  and the image backbone. 
As shown in Tab.~\ref{tab: scale}, comparing rows 1, 4, and 5, we scale the image backbone from ResNet34 to ResNet50 and ResNet101, while comparing rows 3, 4, and 5, we scale the size of the hidden dimension from 125 to 256 and 384.

The ablation results demonstrate that World4Drive exhibits excellent scalability for both image backbone and hidden dimension, which is different from previous methods~\cite{jia2025drivetransformer}. 
In previous methods, increasing hidden dimensions typically yields greater benefits than scaling up the image backbone. 
Our method shows comparable improvements with both scaling strategies. 
It is natural because the extracted latent representations are directly utilized for planning tasks, and both caling strategies can effectively incorporate additional scene information that directly benefits vehicle planning.

\subsection{Qualitative Results}
\begin{figure*}[!t]
    \centering
    \includegraphics[width=1\textwidth]{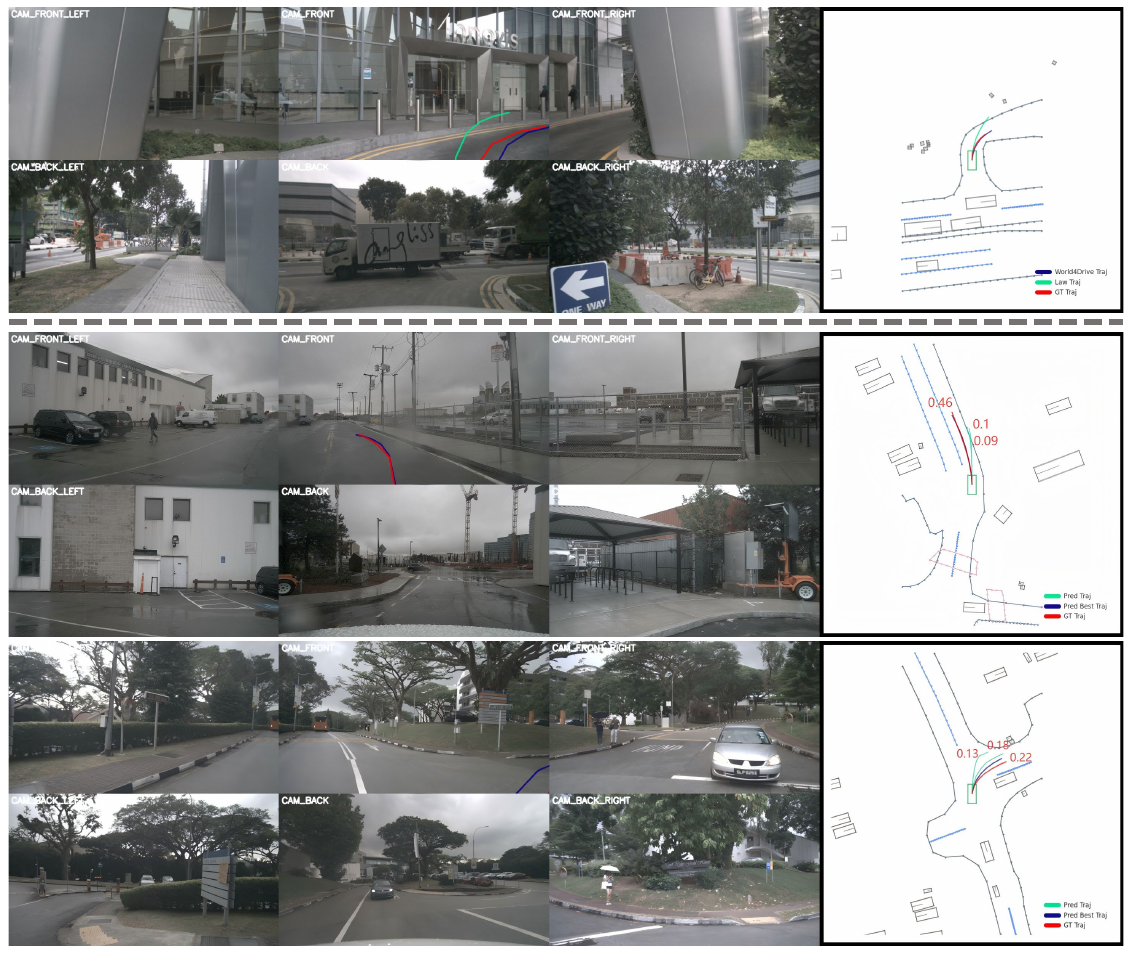}
    \vspace{-2mm}
    \caption{Visualization of World4Drive. Since World4Drive does not predict explicit perception results, we render the ground truth annotations as the perception results.}
    \label{fig: result}
\end{figure*}
In this section, we present the visualization of World4Drive on nuScenes benchmark. 
The qualitative result is shown in Fig.~\ref{fig: result}. 
The upper portion of the visualization demonstrates that World4Drive does safer planning during turning maneuvers compared to LAW.
The lower portion shows that the world model selector effectively selects the most reasonable trajectory from multi-modal planning intentions across diverse scenarios.
Additional visualizations and failure case analyses are provided in the supplementary material.
\section{Conclusion}
In this paper, we present World4Drive, an intention-aware physical latent world model.
World4Drive proposes a novel framework that incorporates driving intentions with a latent world model, innovatively leveraging a latent world model to generate, evaluate, and select multi-modal trajectories.
Specifically, World4Drive proposes a physical world latent encoding module, incorporating the spatial and semantic priors from vision foundation models and aggregating temporal information.
Extensive experiments on nuScenes and NavSim benchmarks demonstrate World4Drive's profound and comprehensive understanding of the physical world, as well as the effectiveness of tightly coupling driving intentions with the latent world model.
{
    \small
    \bibliographystyle{ieeenat_fullname}
    \bibliography{main}
}

\end{document}